\pdfoutput=1
\documentclass[sigconf]{acmart}
\AtBeginDocument{%
  \providecommand\BibTeX{{%
    \normalfont B\kern-0.5em{\scshape i\kern-0.25em b}\kern-0.8em\TeX}}}

\copyrightyear{2022} 
\acmYear{2022} 
\setcopyright{acmcopyright}\acmConference[JCDL '22]{The ACM/IEEE Joint Conference on Digital Libraries in 2022}{June 20--24, 2022}{Cologne, Germany}
\acmBooktitle{The ACM/IEEE Joint Conference on Digital Libraries in 2022 (JCDL '22), June 20--24, 2022, Cologne, Germany}
\acmPrice{15.00}
\acmDOI{10.1145/3529372.3530924}
\acmISBN{978-1-4503-9345-4/22/06}

\usepackage{array, float, multirow}

\hypersetup{
   breaklinks=true,   
   colorlinks=true,   
   pdfusetitle=true,  
}

\newcolumntype{C}[1]{>{\centering\arraybackslash}p{#1}}
\newcolumntype{L}[1]{>{\arraybackslash}p{#1}}

\newcommand{\subject}[1]{\textbf{#1}$^\text{[S]}$}
\newcommand{\object}[1]{\textbf{#1}$^\text{[O]}$}

\newcommand{\subjectAnnotated}[2]{\subject{#1} \textbf{(#2)}}
\newcommand{\predicate}[1]{\emph{#1}$^\text{[P]}$}
\newcommand{\objectAnnotated}[2]{\object{#1} \textbf{(#2)}}

\newcommand{\openie}{OpenIE}
\newcommand{\openieS}{OpenIE6}

\begin{document}

\title[A Library Perspective on Nearly-Unsupervised Information Extraction]{A Library Perspective on Nearly-Unsupervised Information Extraction Workflows in Digital Libraries}

\author{Hermann Kroll}
\email{kroll@ifis.cs.tu-bs.de}
\orcid{0000-0001-9887-9276}
\affiliation{%
  \institution{Institute for Information Systems, TU Braunschweig}
  \streetaddress{Mühlenpfordtstr. 23}
  \city{Braunschweig}
  \state{Lower Saxony}
  \country{Germany}
  \postcode{38106}
}   

\author{Jan Pirklbauer}
\email{j.pirklbauer@tu-bs.de}
\affiliation{%
  \institution{Institute for Information Systems, TU Braunschweig}
  \streetaddress{Mühlenpfordtstr. 23}
  \city{Braunschweig}
  \state{Lower Saxony}
  \country{Germany}
  \postcode{38106}
}   

\author{Florian Plötzky}
\email{ploetzky@ifis.cs.tu-bs.de}
\orcid{0000-0002-4112-3192}
\affiliation{%
  \institution{Institute for Information Systems, TU Braunschweig}
  \streetaddress{Mühlenpfordtstr. 23}
  \city{Braunschweig}
  \state{Lower Saxony}
  \country{Germany}
  \postcode{38106}
}   

\author{Wolf-Tilo Balke}
\email{balke@ifis.cs.tu-bs.de}
\orcid{0000-0002-5443-1215}
\affiliation{%
  \institution{Institute for Information Systems, TU Braunschweig}
  \streetaddress{Mühlenpfordtstr. 23}
  \city{Braunschweig}
  \state{Lower Saxony}
  \country{Germany}
  \postcode{38106}
}   
\renewcommand{\shortauthors}{Kroll et al.}

\begin{abstract}
Information extraction can support novel and effective access paths for digital libraries. 
Nevertheless, designing reliable extraction workflows can be cost-intensive in practice. 
On the one hand, suitable extraction methods rely on domain-specific training data. On the other hand, unsupervised and open extraction methods usually produce not-canonicalized extraction results. 
This paper tackles the question how digital libraries can handle such extractions and if their quality is sufficient in practice. 
We focus on unsupervised extraction workflows by analyzing them in case studies in the domains of encyclopedias (Wikipedia), pharmacy and political sciences.
We report on opportunities and limitations. 
Finally we discuss best practices for unsupervised extraction workflows.
\end{abstract}



\maketitle

\section{Introduction}
Extracting structured information from textual digital library collections enables novel access paths, e.g., answering complex queries over knowledge bases~\cite{auer2007dbpedia,menasha2020jcdllbdworkflow}, providing structured overviews about the latest literature~\cite{auer2019openknowledgeresearchgraph}, or discovering new knowledge~\cite{hristovski2015semmeddblbd}.
However, utilizing information extraction (IE) tools in digital libraries is usually quite cost-intensive which hampers the implementation in practice. 
On the one hand, extraction methods usually rely on supervision, i.e., ten  thousands of examples must be given for training suitable extraction models~\cite{weikum2021machine}.
On the other hand, utilizing the latest natural language processing (NLP) tools in productive pipelines requires high expertise and computational resources.  

In addition to supervised IE, Open IE methods (\openie) have been developed to work out-of-the box without additional domain-specific training~\cite{manning2014stanford,kolluru2020openie6}.
But why aren't they used broadly in digital library applications?
The reason is that \openie{} generates non-canonicalized (not normalized) results, i.e., several extractions describing the same piece of information may be structured in completely different ways (synonymous relations, paraphrased information, etc.). 
But such non-canonicalized results are generally not helpful in practice, because a clear relation and entity semantics like in supervised extraction workflows is vital for information management and query processing.
Since the lack of clear semantics has been recognized as a major issue, cleaning and canonicalization methods have been investigated to better handle such extractions~\cite{vashishth2018cesi}.
Still are they ready for application in digital libraries?

\textit{In this paper case studies are used to find out how suitable nearly-unsupervised methods are to design reliable extraction workflows. In particular  we analyze extraction and cleaning methods from the perspective of a digital library by assessing the required expertise, domain knowledge, computational costs and result quality.}

Therefore we selected our toolbox for a nearly-unsupervised extraction from text published in last year's JCDL~\cite{kroll2021toolbox}.
The toolbox contains interfaces to the latest named entity recognition (NER) and open information extraction methods. 
In addition, it includes cleaning and canonicalization methods to handle noisy extractions by utilizing domain-specific information. 
Our corresponding paper~\cite{kroll2021toolbox} advertises the toolbox to considerably decrease the need for supervision and to be transferable across domains, nevertheless it comes with several limitations:
\begin{enumerate}
    \item Although we did report on the extraction quality (good precision, low recall), we did not report on the costs of applying the toolbox, i.e., how much expertise and computational costs are required for a reliable workflow. 
    \item We applied the toolbox only in the biomedical domain, which lessens the generalizability of our findings.
    \item Moreover, we did not report what is technically and conceptually missing in such extraction workflows.
\end{enumerate}
In this paper we address the previous issues by analyzing the toolbox application in three distinct real-world settings from a library perspective:
1. We extracted knowledge about scientists from the online encyclopedia Wikipedia (controlled vocabularies, descriptive writing). 
2. We applied the toolbox to the pharmaceutical domain (controlled vocabularies, entity-centric knowledge) in cooperation with the specialized information service for pharmacy (\url{www.pubpharm.de}). 
3. We applied the toolbox in political sciences (open vocabulary, topic/event-centric knowledge) in cooperation with the specialized information service for political sciences~\cite{SchardelmannOtto2018pollux} (\url{www.pollux-fid.de}).
For Pharmacy and Political Sciences, we recruited associated domain experts for expertise in the evaluation. We performed these three case studies to answer the following questions:
\begin{enumerate}
\item \textit{How much expertise and effort is required to apply nearly-unsupervised extractions across different domains?}
\item \textit{How generalizable are these state-of-the-art extraction methods and particularly, how useful are the extraction results?}
\item \textit{What is missing towards a comprehensive information extraction from texts, e.g., for retaining the original information?}
\end{enumerate}

\section{Study Objectives}
In the following we briefly summarize the nearly-unsupervised extraction toolbox, raise research questions for our case studies, and explain why we selected the three domains here.
Our main objective is to analyze unsupervised extraction workflows from a digital library perspective.

\subsection{Overview of the Toolbox}
The extraction toolbox covers three common IE areas: entity detection, information extraction and canonicalization. 
We shared our toolbox as open-source software and made it publicly available\footnote{\url{https://github.com/HermannKroll/KGExtractionToolbox}}.
We focus on this toolbox because it proposed an eased and nearly-unsupervised extraction workflow by integrating latest unsupervised extraction plus suitable cleaning methods.

\textit{Entity Detection.} 
The toolbox integrates interfaces to one of the latest NER tools, Stanford Stanza~\cite{qi2020stanza}. 
Stanza is capable of detecting 18 general purpose entity types like \textit{persons}, \textit{organizations}, \textit{countries}, and \textit{dates} in texts; See \cite{qi2020stanza} for a complete overview. 
In addition, the toolbox supports the linking of custom entity vocabularies via a dictionary-based lookup method.
The entity linker supports an abbreviation resolution and handling of short homonymous terms (link if the entity is mentioned with a longer mention in the text).

\textit{Information Extraction.}
The toolbox integrates implements interfaces to \openie{} methods, Stanford CoreNLP~\cite{manning2014stanford} and \openieS~\cite{kolluru2020openie6}. 
Besides, the toolbox includes a self-developed path-based extraction method named PathIE.
PathIE extracts statements between entities in a sentence if connected in the grammatical structure via verb phrases or custom keywords (e.g., treatment, inhibition, award, and member of) that can be specified beforehand. 
The \openie{} methods work entirely without entity information, whereas the PathIE requires entity annotations as starting points.

\textit{Cleaning and Canonicalization.}
\openie{} and PathIE may produce non-helpful and non-canonicalized outputs, i.e., synonymous noun and verb phrases that describe the same information.
The toolbox supports canonicalizing and filtering such outputs automatically. 
First, extracted noun phrases can be filtered by entity annotations, i.e., only noun phrases that include relevant entities are kept. 
Here three different filters are supported to filter noun phrases: exact (noun phrase matches an entity), partial (noun phrase partially includes an entity), and no filter (keep original noun phrase).

Second, an iterative cleaning algorithm is integrated that can canonicalize synonymous verb phrases to precise relations, e.g., \textit{birthplace} or \textit{place of birth} to \textit{born in}.
Therefore, users can export statistics about the non-canonicalized verb phrases and build a so-called relation vocabulary.
Each entry of this vocabulary is a relation consisting of a name and a set of synonymous. 
The toolbox utilizes this vocabulary to automatically map synonymous verb phrases to precise relations.
Word embeddings are supported in the canonicalization procedure to bypass an exhausting editing of the relation vocabulary.
The central idea of word embeddings is that words with a similar context appear close in the vector space~\cite{mikolov2013efficient}.
The word embedding is then used to automatically map a new verb phrase to the closest match (most similar) in the vocabulary. 
Relation type constraints can then be used to filter the extractions further, i.e., a relation type constraint describes which entity types are allowed as subjects and objects.
For example, \textit{born in} can be defined as a relation between \textit{persons} and \textit{countries}.
Other extractions that hurt these constraints are then removed.
We did already report on some challenges of \openie{} extractions, especially on handling noun phrases~\cite{kroll2021discoopenie}.
In contrast to our previous works, this work analysis the complete workflow in three domains from a library perspective.

\subsection{Study Goals}
The study goals concern three concrete areas of study: 1. application costs, 2. generalizability, and 3. limitations for a comprehensive IE.
However answering these questions on a purely quantitative level is challenging, e.g., how can the costs be measured?
That is why we report our findings as a mixture of quantitative measures (e.g., time spent and runtimes) and qualitative observations (what works well and what not).
We define evaluation criteria for all of the three aspects in the following.

\textit{Application Costs.} We understand everything necessary to implement a workflow with the toolbox as \textit{application costs}. 
We estimate the application costs in terms of
\begin{description}
    \item[Data Preparation:]  transforming data into toolbox formats (e.g., JSON), working with toolbox outputs (TSV/JSON)
    \item[Implementation:] computational costs (runtime and space), scalability, executed steps, effort to choose parameters, encountered issues
    \item[Domain Knowledge:] entity and relation vocabulary design, required knowledge for canonicalization
\end{description}

\textit{Generalizability.} In short, how well are the proposed methods generalizable across domains and how useful are the results?

\begin{description}
    \item[Extraction quality:] benchmarks (precision and recall), observations, extraction limitations
    \item[Usefulness:] relevance of statements (e.g., non-obvious statements), domain insights, helpfulness for domain experts, usefulness in applications
\end{description}

Information, originally connected in coherent written texts, might be broken into not helpful pieces in the end. 
For a good example, consider a drug-disease treatment:
Here context information like the dose or treatment duration, which could give more information about the statement's validity~\cite{kroll2020tpdl}, might get lost.
We refer to such information as the \textbf{context} of statements, e.g., the surrounding scope in which a statement is valid.
In addition, the connection between statements might get lost too, e.g., an assumption might lead to a conclusion. 
We call this the \textbf{coherence of statements}.
They are crucial for real-world applications, but are they yet considered?

\textit{On Context and Coherence.} Contexts affect the validity of statements and coherence describes how statements belong together.
We evaluate the following criteria:

\begin{description}
    \item[Contexts:] relevance of contexts, which kind of information requires context, how does the context affect the validity of extracted statements, what must be done to retain context
    \item[Coherence:] complex information that is broken into pieces, which kind of information is broken down, what are the subsequent problems with such a decomposition 
\end{description}

\begin{table}
    \centering
    \caption{The number of documents and sentences is reported for each collection and sample.}
    \label{tab:dataset_statistics}
    \begin{tabular}{@{} |l|l|cc|  @{}}
        \toprule
          Collection & Size  & \multicolumn{2}{c|}{Sample} \\
         & & \emph{\#Documents} & \emph{\#Sentences}   \\ 
          \midrule
          English Wikipedia & 6.3M & 2,373 & 74.5k  \\
          PubMed & 33M & 10k & 87.1k  \\
          Political Sciences & 1.7M & 10k & 66.9k  \\
        \bottomrule
    \end{tabular}
\end{table}

\subsection{Case Study Selection}
We applied the toolbox in three different domains to generalize the findings in this paper. 
Here we focused on natural language texts written in the English language. 
We describe the domains and their characteristics in the following.
Statistics about the used data sets and samples are listed in Table~\ref{tab:dataset_statistics}.

\paragraph{Wikipedia.} 
A prime example of an encyclopedia is the free and collaborative Wikipedia.
Encyclopedic texts should be written in a descriptive and objective language, i.e., wording and framing should not play any role.
Wikipedia captures knowledge about certain items (persons, locations, events, etc.), in our understanding, entities. 
Here controlled ontologies about entities and relations are available; See Wikidata~\cite{vrandevcic2014wikidata} as a good example.
However Wikipedia texts also tend to include very long and complex sentences.
For this case study we focus on knowledge about famous fictional and non-fictional scientists (about 2.4k scientists with an English Wikipedia article and Wikidata entry).
This case study was selected because sentences are written objectively and controlled vocabularies are available for usage.

\paragraph{Pharmaceutical Domain.}
The pharmaceutical domain focuses on entity-centric knowledge, i.e., statements about entities such as drugs, diseases, treatments, and side effects.
Many vocabularies and ontologies are curated to describe relevant biomedical entities, e.g., the National Library of Medicine (NLM) maintains the so-called Medical Subject Headings (MeSH)\footnote{\url{https://meshb.nlm.nih.gov/search}}. 
These headings are entities with descriptions, ontological relations (subclasses), and suitable synonyms.
In this paper we select a subset of the most comprehensive biomedical collection, the NLM Medline collection\footnote{\url{https://www.nlm.nih.gov/medline/medline_overview.html}}. 
Medline includes around 33 million publications with metadata (title, abstracts, keywords, authors, publication information, etc.).
The specialized information service for pharmacy was interested in statements about drugs. 
That is why we selected a PubMed subset that contains drugs. 
Therefore, we applied the entity linking step to all Medline abstracts (Dec. 2021) and then randomly picked a subset of 10k abstracts that include at least one drug mention.

\paragraph{Political Sciences.} 
The political sciences domain encompasses a diverse range of content, e.g., publications about topics and events, debates, news, and political analyses.
Due to its diversity this domain does not have extensive curated vocabularies and ontologies available. 
We argue that entity subsets of knowledge bases like Wikidata~\cite{vrandevcic2014wikidata} or DBpedia~\cite{auer2007dbpedia} might be good starting points to derive some entity vocabularies regarding persons, events, locations, and more.
Still Wikidata and DBpedia are built as general-purpose knowledge bases and are thus not focused on political sciences (in contrast to MeSH for the biomedical domain).
Nevertheless they might be helpful to analyze texts in political sciences and that is why we analyze them for a practical application here. 
In addition, descriptions of entities in political sciences tend to be subjective, i.e., they depend on different viewpoints and schools of thought. 
For example, the accession of Crimea to Russia in 2014 was a highly discussed topic whether this event could be seen as peaceful secession or as an annexation.
In contrast to objective and entity-centric statements in biomedicine, political sciences are far more based on the wording and framing of certain events. 
This case study analyzes how far IE methods can bring structure into these texts and where these methods fail.
The specialized information service for political sciences (Pollux) provided us with around three million publications (around 1.3 million English abstracts).
Our case study is based on a random sample of 10k abstracts selected from the English subset. 
In addition, domain experts manually selected five abstracts due to their focus on the diverse topics of the EU, philosophy, international relations, and parliamentarism.

\section{Case Studies}

\begin{table*}
    \centering
    \caption{Extraction statistics for all three domains:
    Sentences (number, percentage of complex sentences, number of sentences with at least two entities mentions), Entity Detection (number of Stanza NER and dictionary-based entity linking annotations), \openieS{} (percentage of complex subjects and objects, number of extractions computed by the different entity filters [no, partial, exact, subject]) and PathIE (number of extractions).} \label{tab:extraction_statistics}
    \begin{tabular}{|c|c|c|c|c|c|c|c|c|c|c|c|c|c|}
    \toprule
    & \multicolumn{3}{c|}{\textbf{Sentences}} & \multicolumn{2}{c|}{\textbf{Entity Det.}} & \multicolumn{6}{c|}{\textbf{\openieS}} & \textbf{PathIE} \\
    &\emph{\#Sent.} & \emph{Compl.} & \emph{\#w2E} & \emph{\#NER} & \emph{\#EL} & \emph{C. Subjs.} & \emph{C. Objs.} & \emph{\#No EF} & \emph{\#Part. EF} & \emph{\#Exact EF} & \emph{\#Subj. EF} & \emph{\#Extr.} \\
    \midrule
    Wikipedia & 74.5k & 92.7\% & 50.3k & 155.0k & 113.2k & 16.2\% & 74.5\% & 177.1k & 317.8k & 2.9k & 80.9k & 1.3M  \\ 
    Pharmacy & 87.1k & 92.2\% &  47.4k & - & 232.5k & 37.8\% & 72.1\% & 207.6k & 88.0k & 291 & 151.0k & 430.8k \\
    Pol. Sci. & 66.9k & 93.2\% & 17.6k & 80.0k & 3.7k &  32.0\% & 74.3\% & 147.2k & 28.6k & 128 & 7.3k & - \\
    \bottomrule
    \end{tabular}
\end{table*}

For our case studies we developed scripts, produced intermediate results, and implemented some improvements for the toolbox.
The details, used data and produced results of every case study can be found in our evaluation scripts on GitHub (see the Toolbox GitHub Repository).
We included a Readme file to document the following case studies.
All of our experiments and time measurements were performed on our server, having two Intel Xeon E5-2687W (\@3,1GHz, eight cores, 16 threads), 377GB of DDR3 main memory, one Nvidia 1080 TI GTX GPU, and SSDs as storage.

\subsection{Wikipedia Case Study}
This first case study was based on 2.3k English Wikipedia full-text articles about scientists.
The conversion of Wikipedia articles was simple: We downloaded the available English Wikipedia dump (Dec. 2021), used the WikiExtractor~\cite{Wikiextractor2015} to retrieve plain texts, and filtered these texts by our scientist's criteria (title must be about a scientist of Wikidata).
Next we developed a Python script to transform the plain texts into a JSON format for the toolbox.
The data transformations took half a person-day. 

\begin{table*}
  \centering
  \caption{\openieS{} example extractions from the Wikipedia article of Albert Einstein. On the left the corresponding entity filter is shown (subject, partial and exact). Subject$^\text{[S]}$, predicate$^\text{[P]}$ and object$^\text{[O]}$ are highlighted respectively.}
  \label{tab:wikipedia_extractions}
  \begin{tabular}{@{}|l|l|lL{0.8\textwidth}|@{}}
     \toprule
         \multirow{9}{*}{\rotatebox[origin=c]{90}{\textbf{Wikipedia}}} &
         
        \parbox[t]{2mm}{\multirow{3}{*}{\rotatebox[origin=c]{90}{\textbf{Exact}}}} &
         
         E1.1 & In 1933, while \subjectAnnotated{Einstein}{Person} \predicate{was visiting} \objectAnnotated{the United States}{Country}, [...] \\ 
         
        && E1.2 & On 30 April 1905, Einstein completed his thesis, with \subjectAnnotated{Alfred Kleiner}{Person}, \predicate{[be] Professor} of \objectAnnotated{Experimental Physics}{ORG}, serving as "pro-forma" advisor. \\ 
         \cmidrule{2-4}
         &\parbox[t]{2mm}{\multirow{3}{*}{\rotatebox[origin=c]{90}{\textbf{Partial}}}}&
        E2.1. & In a German-language letter to \objectAnnotated{philosopher}{Profession} Eric Gutkind, dated 3 January 1954, \subjectAnnotated{Einstein}{Person} \predicate{wrote}: [...]\\
        && E2.2 & \subjectAnnotated{Einstein}{Person}  \predicate{was elected} a Foreign Member of \objectAnnotated{the Royal Society}{Org} (ForMemRS) in 1921. \\
         \cmidrule{2-4}
        
         &\parbox[t]{2mm}{\multirow{3}{*}{\rotatebox[origin=c]{90}{\textbf{Subject}}}}&
        
         E3.1 & During an address to Caltech's students, \subjectAnnotated{Einstein}{Person} \predicate{noted} \object{that science was often inclined to do more harm than good}.\\
         
         && E3.2 & \subjectAnnotated{Einstein}{Person} \predicate{started teaching} himself \object{calculus at 12}, and as a 14-year-old [...] \\
     \bottomrule
  \end{tabular}
\end{table*}

\textit{Entity Linking.}
In this case study we focused on statements about scientists such as works, scientific organizations, and degrees.
Therefore, we performed entity linking to identify these concepts and use them to filter the extraction outputs.
We derived corresponding entity vocabularies from Wikidata by utilizing the official SPARQL endpoint. 
We retrieved vocabularies by asking for English labels and alternative labels for the following entity types: \emph{Academia of Sciences}, \emph{Awards}, \emph{Countries}, \emph{Doctoral Degrees}, \emph{Religions} and \emph{Irreligions}, \emph{Scientists}, \emph{Professional Societies},  \emph{Scientific Societies} and \emph{Universities}.
We adjusted the SPARQL queries to directly download the vocabularies as TSV files in the toolbox format.

A first look over this entity vocabulary revealed some misleading labels (e.g., the, he, she, and, or), which we removed. 
We applied the dictionary-based entity linker utilizing our vocabulary on the articles.
The linker yielded many erroneously linked entities due to very ambiguous labels in the dictionary, e.g., the mentions \textit{doctor}, \textit{atom} and \textit{observation} were linked to fictional characters which are scientists regarding the Wikidata ontology. 
Next synonyms like \textit{Einstein} were erroneously linked when talking about his family or talking about the term \textit{Einstein} in the sense of \textit{genius}. 
The linker also ignored pronouns completely, i.e., no coreference resolution was applied. 
Especially in Wikipedia articles, pronouns are often used.
In addition, we executed Stanford Stanza to recognize general-purpose entity types like dates or organizations.
We found short entity names to be too ambiguous. 
That is why we removed all detected entities with less than five characters.
This step yielded 155k Stanza NER mentions and 113.2k dictionary-based entity links.

\textit{Information Extraction.}
We applied the \openieS{} method and the entity filter methods (no filter, partial, exact). 
We obtained 117.1k (no filter), 317.8k (partial) and 2.9k (exact) extractions.
Note that statements can be duplicated for the partial filter if multiple entities are included within the same noun phrase.
We exported 100 results for each filter randomly and analyzed them.
In the following we report on some examples of good and bad extractions.

Some interesting results about Albert Einstein are listed in Table~\ref{tab:wikipedia_extractions}.
\openieS{} produced correct and helpful extractions when sentences were short and simple (no nested structure, no relative clauses, etc.).
When sentences became longer, the tool yielded short subjects but long and complex objects, e.g., a whole subordinate clause like \textit{that science was often inclined to do more harm than good}.
See E3.1 in Table~\ref{tab:wikipedia_extractions}.

We developed a short script to quantify them to better understand how many sentences, subjects, and objects were complex.  
Therefore, we formulated regular expressions to check if a sentence contained multiple clauses split by punctuation (,|;|:), or words (and|or|that|thus| hence|because|due|etc.).
We counted sentences, subjects, and objects as complex if they matched one of these regular expressions. 
In addition, if a sentence was denoted as complex and the extracted noun phrase was larger than 50\% (character count) of the sentence or it contained words like (by|at|for|etc.), we considered it complex.
For our sample, 92.7\% of the sentences, 16.2\% of subjects, and 74.5\% of objects were classified as complex.
We iterated over these classifications to verify the filter criteria. 

\textit{Partial Entity Filter.} This filter yielded problematic results because much information was lost, e.g., a whole subordinate clause was broken down to a single entity regardless of where the entity appeared in this clause.
In some cases, this filtering completely altered the sentence's original information; See E2.2 for a good example.
Here the extraction \textit{Einstein was elected the Royal Society} was nonsense because \textit{Foreign Member} was filtered out.
In E2.1, the extracted statement missed that the \textit{philosopher} was \textit{Eric Gutkind}, and thus lost relevant information.

\textit{Exact Entity Filter.} The exact filter was very restrictive because the number of extractions was reduced from 117.9k to 2.9k.
However the extraction seemed to have good quality.
In E1.1, the extraction \textit{Einstein was visiting the US} was correct, but the context about the year \textit{1933} was lost.
Extraction E1.2 showed that \openieS{} was capable of extracting implicit statements like \textit{be Professor of}. 
Again, the surrounding context about the year and Einstein was lost.
Other extractions showed that a coreference resolution would be beneficial to resolve mentions like \textit{his}, \textit{in the same article}, and, \textit{these models}.

We observed many complex object phrases (74.5\% in sum).
These complex phrases contained more information than a single entity. 
Filtering them led to many wrongly extracted statements.
In contrast, subject phrases were often simple and might stand for a single entity (only 16.2\% are complex).
Due to these observations, we developed a subject entity filter, i.e., only subjects had to match entities directly. 
The idea was to identify subjects as precise entities and keep object phrases in their original form to retain all information.

\textit{Subject Entity Filter}. 
This filter worked as expected: 
In E3.1 and E3.2, the subject was identified as the Person \textit{Einstein} whereas the original information was kept in the object phrase.
This filtering allowed us to generate a structured overview about Albert Einstein, for example.

In addition to \openieS, we investigated how useful PathIE is to extract relations between the relevant entity types such as scientists and awards. 
PathIE allowed us to specify keywords that can indicate a relation.
In a first attempt, we applied PathIE with a small relation vocabulary of Wikidata.
We exported the English labels and alternative labels of eleven Wikidata properties that describe the relations between the given entity types: academic degree, award received, date of birth, date of death, field of work, member of, native language, occupation, religion, and writing language.

We exported 100 randomly selected PathIE extractions for evaluation.
When several entities were detected in long and nested sentences, PathIE yielded many wrong extractions because the corresponding entities were connected via some verb phrases, e.g., \textit{Einstein return Zurich} from \textit{Einstein visited relatives in Germany while Maric returned to Zurich} or \textit{Written languages write Leningrad}.
Filtering these extractions by entity types like (Person, Date) or (Person, Award) revealed more helpful extractions, e.g., \textit{Einstein win Nobel Prize} from \textit{Einstein received news that he had won the Nobel Prize in November}.

However we encountered severe entity linking issues when analyzing the cleaned \openieS{} and PathIE extractions.
On the one hand, ambiguous terms were linked wrongly.
On the other hand, fragments of a text span were linked against an entity although the whole text span referred to a single entity, e.g., only linking \textit{Albert Einstein} in the text mention \textit{Albert Einstein's Theory of Relativity was published in 1916}.
These issues directly affected the extraction quality.
We stopped the extraction part at this point.

\textit{Canonicalization.}
We used our small relation vocabulary to canonicalize the extractions.
This procedure did work out for PathIE because it directly extracted the vocabulary entries from the texts. 
For example we could retrieve a list of statements that indicate an \textit{award received} relation.
However further cleaning was required to obtain \textit{award received} relations between persons and awards.
We analyzed 100 entries for this relation.
Although some extraction were correct, 60 of 100 extractions had linked awards that were not helpful, e.g., \textit{awards}, \textit{doctor}, \textit{medal}, \textit{president} and \textit{master}.
The remaining 40 extractions displayed six wrongly identified persons.
However the remaining 34 extractions seemed to be plausible, although some information was missed, like the \textit{Nobel prize's} category.

In contrast, the canonicalization procedure did not work for \openieS{} extractions.
The reason was that the extracted verb phrases did not appear directly in the vocabulary. 
Thus we used a pre-trained English Wikipedia word embedding from fasttext\footnote{\url{https://fasttext.cc/docs/en/pretrained-vectors.html}} to find similar matches in the relation vocabulary. 
We adjusted the cleaning parameters (how similar terms must be and how often terms must occur) and canonicalized the \openieS{} verb phrases.
However most verb phrases were mapped wrongly because the vocabulary was relatively small, e.g., \textit{divorce} was mapped to \textit{date of death} because it was the closest match.

We then derived a list of 120 Wikidata properties that involved persons (ignoring usernames and identifiers) to find more matches. 
We repeated the canonicalization and analyzed 100 extractions obtained by the subject entity filter because it retrieved the most helpful results in the previous step. 

Most of the canonicalized verb phrases were mapped incorrectly, e.g., mapping \textit{start teach} to \textit{educated at} or \textit{begin} to \textit{death of place} was wrong.
For a positive example, the verb phrase \textit{publish} was mapped to the relation \textit{notable work} and \textit{write} to \textit{author}, e.g., \textit{Galileo publish ($\mapsto$ notable work) Dialogue Concerning the Two Chief World Systems}.
Although this relation was correct for some fewer extractions, most of these mappings were problematic, e.g., \textit{Einstein publish ($\mapsto$ notable work) his own articles describing the model among them.} 
Here the object phrase did not contain a notable work in the sense of how we would understand it.
In summary, the canonicalization procedure had many problems for \openieS{} extractions.
The main issue was that the canonicalization procedure only considered the verb phrase and not the surrounding context in a sentence. 
But this surrounding context is essential to determine the relation.
In addition, the relation vocabulary obtained from Wikidata might be insufficient because it did not contain verb phrases as we would expect them.
Wikidata describes relations by using substantives and nouns, e.g., notable work of, notable work by, notably created by for the relation \textit{notable work}. 

\textit{Application Costs.}
We spent much of our time understanding the Wikidata ontology and formulating suitable SPARQL queries to retrieve the utilized vocabularies. 
The corresponding vocabularies could be exported directly from Wikidata and did not need transformations besides concatenation of files. 
We formulated several SQL queries to analyze, clean, and filter entity annotations and extractions in the toolbox's underlying database.
In summary, three persons performed this case study within three person-days.

\textit{Generalizability.}
We had a close look at existing Wikipedia relation extraction benchmarks for evaluation.
Unfortunately, these benchmarks are often built distantly supervised, i.e., if two entities appear in a sentence, and both entities have a relation in a knowledge base, then this relation is the class that must be predicted for this sentence.
In other words, the relation does not have to appear within the sentence. 
Furthermore these benchmarks often require domain knowledge, e.g., if a football player started his career at a sports team, then the football player \textit{played for} this team.
This additional knowledge is typically not included in \openie{} methods. 
\openie{} extracts statements based on grammatical patterns in a sentence: 
For the previous example, the tool would extract that the football player started his career at the sports team, but not that he also played for the team.
That is why we did not evaluate the extraction tool on existing benchmarks because we expected the quality to be low by design.
Moreover, mapping verb phrases to precise relations would also be challenging.
In contrast, we wanted to understand how useful the results were for practical applications.

First, an improved entity linking would have solved several issues in our case study. 
Next the handling of complex noun phrases was an issue: 
Although the exact entity filter was too restrictive, it resulted in suitable extractions.
The partial entity filter messed up the original information and was thus not helpful. 
\openieS{} and the subject entity filter allowed us to retrieve a list of actions performed by Albert Einstein, for example.
However this filtering did not yield a canonicalized knowledge base by design.
Our case study has shown that PathIE could extract relations between scientists and awards.
Although we could not evaluate the quality in rough numbers, we spent three person-days designing a possible extraction workflow.
Here the toolbox allowed us to retrieve such semi-structured information in an acceptable amount of time. 

\textit{What is missing.} 
Handling of complex noun phrases was a significant issue:
On the one hand, the decisive context was lost if phrases were broken down into small entities.
On the other hand, if phrases were retained in their original form, context was kept, but the canonicalization remained unclear.
To the best of our knowledge there is no out-of-the-box solution that will solve these issues.

\subsection{Pharmaceutical Case Study}
We applied the toolbox to a subset of the biomedical Medline collection for our second case study. 
The PubMed Medline is available in different formats, among other things, in the PubTator format which is supported by the toolbox. 
We downloaded the document abstracts from the PubTator Service~\cite{Wei2013pubtator}.

\begin{table*}
    \centering
    \caption{PubMed PathIE example extractions. On the left the canonicalized relation is annotated. } \label{tab:pharma_extractions}
    \begin{tabular}{@{}|l|l|lL{0.8\textwidth}|@{}}
    \toprule
      \multirow{11}{*}{\rotatebox[origin=c]{90}{\textbf{Pharmacy}}} &
         \parbox[t]{2mm}{\multirow{4}{*}{\rotatebox[origin=c]{90}{\textbf{Treats}}}}&
         
         P1.1 & 
         We tested whether short-term, low-dose \predicate{treatment} with the fluvastatin and \subjectAnnotated{valsartan}{drug} combination could improve impaired arterial wall characteristics in type 1 \objectAnnotated{diabetes mellitus}{disease} patients. \\
    
    && P1.2. & We encountered two cases of cerebellar \objectAnnotated{hemorrhage}{Disease} in patients \predicate{treated} with \subjectAnnotated{edoxaban}{Drug} for PVT after hepatobiliary surgery during the past 2 years. \\

         \cmidrule{2-4}
         & \parbox[t]{2mm}{\multirow{3}{*}{\rotatebox[origin=c]{90}{\textbf{Inhibits}}}}& 
         
         P2.1 & \subjectAnnotated{Anthraquinone}{Drug} derivative emodin inhibits tumor-associated angiogenesis through \predicate{inhibition} of \objectAnnotated{extracellular signal-regulated kinase 1}{Gene}/2 phosphorylation. \\
         
         && P2.2 & Impact of \subjectAnnotated{aspirin}{Drug} on the gastrointestinal-sparing effects of \objectAnnotated{cyclooxygenase-2}{Gene} \predicate{inhibitors}.\\
         
         \cmidrule{2-4}
         & \parbox[t]{2mm}{\multirow{3}{*}{\rotatebox[origin=c]{90}{\textbf{Induces}}}}&
         
         P3.1 & \objectAnnotated{Hyperglycemia}{Disease}-\predicate{induced} mitochondrial dysfunction plays a key role in the pathogenesis of diabetic \subjectAnnotated{cardiomyopathy}{Disease}. \\
         
         && P3.2 & Conclusions H. pylori \subjectAnnotated{Infection}{Disease} appears to \predicate{cause} decreases in \objectAnnotated{Vitamin B12}{Excipient}[...].\\
    \bottomrule
    \end{tabular}
\end{table*}

\textit{Entity Linking.} 
We utilized existing entity annotations (diseases, genes, and species) from the PubTator Central service.
In addition, we selected subsets of MeSH (diseases, methods, dosage forms), ChEMBL~\cite{mendez2018chembl} (drugs and chemicals), and Wikidata (plant families) to derive suitable entity vocabularies.
We developed scripts that retrieved relevant entries from these vocabularies.
This step required us to export relevant entries from XML and CSV files into TSV files.

We then applied the entity linker and analyzed the results by going through the most frequent annotations.
Our first attempt yielded frequently, but obviously wrongly linked words such as \textit{horse}, \textit{target}, \textit{compound}, \textit{monitor}, and \textit{iris}.
These words were derived from ChEMBL because they were trade names for drugs.
We found such trade names to be very ambiguous and removed them. 
But we also found annotations such as \textit{major}, \textit{solution}, \textit{relief}, \textit{cares}, \textit{aim}, and \textit{advances}. 
We went through the 500 most tagged entity annotations to remove such words by building a list of ignored words.
We repeated the entity linking by ignoring these words and computed 232.5k entity mentions.
We did not apply Stanford Stanza NER here because we were interested in biomedical entities.

\textit{Information Extraction.}
The domain experts were interested in statements between entities. 
That is why we applied \openieS{} and analyzed the partial and exact entity filter.
\openieS{} extracted 207.6k extractions and filtering them yielded 88k (partial) and 291 (exact) extractions.
An analysis of the extractions showed that 92.2\% of sentences, 37.8\% of subjects, and 72.1\% of objects were complex.
The exact entity filter was too restrictive and not helpful because the remaining extractions were too few for a practical application.

\textit{Partial Entity Filter.}
A closer look at 100 randomly sampled extractions indicated that many noun phrases were complex again.
The partial entity filter mixed up the original sentence information by filtering out the important information.
For example consider the following sentence: \textit{Inhibition of P53-MDM2 interaction stabilizes P53 protein and activates P53 pathway}.
Here the partial entity filter extracts the statement: (\textit{MDM2}, \textit{stabilizes}, \textit{protein}). 
This statement mixed up the original information.
Our analysis showed that the vast majority of filtered extractions were incorrect. 
In addition, \openieS{} is focused on verb phrases to extract statements (here \textit{stabilizes}).

However many relevant statements are expressed by using special keywords, e.g. \textit{treatment}, \textit{inhibition}, \textit{side effect}, and \textit{metabolism}. 
That means that these \openie{} methods will usually not extract a statement from clauses like \textit{metformin therapy in diabetic patients} by design.
A similar observation was already made in the original toolbox paper, where \openie{} methods' recall was clearly behind supervised methods (5.8\% vs. 86.2\% and  6.2\% vs. 75.9\% on biomedical benchmarks)~\cite{kroll2021toolbox}.
Supervised extraction methods would engage this problem by learning typical patterns of how a treatment can be expressed within a sentence. 

To integrate such specialized keywords in the extraction process, we applied the recall-oriented PathIE method. 
In the previous example, the entities \textit{metformin} and \textit{diabetic patients} are connected via the keyword \textit{therapy}. 
In this way PathIE extracted a helpful statement. 
However we had to build a relation vocabulary to define these specialized keywords. 
In cooperation with domain experts, we built such a vocabulary by incrementally extracting statements with PathIE, looking at extractions and example sentences to find out what we were missing.
In sum, we had three two-hour sessions to build the final relation vocabulary. 
The final PathIE step yielded 430.8k extractions and took two minutes to complete.
Some interesting results are listed in Table~\ref{tab:pharma_extractions}.
We then iterated over a sample of 100 of these extractions.

PathIE was capable of extracting statements from long and nested sentences, e.g., a treatment statement in P1.1. in Table~\ref{tab:pharma_extractions}.
However we also encountered several issues with PathIE.
If a sentence contains information about treatments' side effects (also linked as diseases), PathIE extracted them wrongly as the treated condition (See P1.2). 
A similar problem occurred when a drug therapy was used to treat two diseases simultaneously. 
Here PathIE yielded six statements (three  mirrored): two therapy statements about the drug and each disease, and one therapy statement between both diseases, which is wrong.
In example P2.2, PathIE failed to recognize that aspirin \textit{effects} the inhibitors and is not an inhibitor  itself.

A second problem was the direction of extracted relations:
A \textit{treats} relation could be defined as a relation between \textit{drugs} and \textit{diseases}.
If a relation has precise and unique entity types, then an entity type filter removes all other, and possibly wrong, extractions. 
Suppose that a disease causes another disease (think about a disease that causes severe effects).
In that case, PathIE would extract both directions: (a causes b) and (b causes a).
For example PathIE would extract two statements from \textit{myocardial damage caused by ischemia-reperfusion}.
Here an entity type filter did not solve the problem because both entities have the type \textit{disease}. 

Third, in situations with several entities and clauses within one sentence, PathIE seemed to mess up the original information and extracted wrong statements, e.g., see P3.1, where hyperglycemia did not induce cardiomyopathy.
In summary, PathIE could extract statements from complex sentences, but a cleaning step had to be applied afterward to achieve acceptable quality.

\textit{Canonicalization.}
We exported the database statistics for PathIE.
We carefully read the extracted verb phrases in cooperation with two domain experts. 
Verb phrases such as \textit{treats}, \textit{prevents} and \textit{cares} point towards a \textit{treats} relation, which we included into our relation vocabulary.
Phrases such as \textit{inhibits} and \textit{down regulates} may stand for a \textit{inhibits} relation. 
To find more synonyms automatically, we used a Biomedical Word Embedding~\cite{zhang2019biowordvec} that we used in the toolbox paper before.
Following this procedure, we defined eight relations with 30 synonyms. 
We repeated the procedure five times and derived a relation vocabulary of 60 entries. 
The relation vocabulary was a mixture of verb phrases and keywords that indicated a relation in the text.
In sum, we had six sessions of two hours each to build the final relation vocabulary. 

However we noticed that PathIE extractions were problematic when not filtered.
Relations like \textit{treats} and \textit{inhibits} also include entity types that we had not expected, e.g., two diseases in treats.
We formulated entity type constraints for eight relations to remove such problematic statements.
The relations \textit{treats} and \textit{inhibits} looked more helpful because they only contained relevant entity types. 
We tried to filter relations like \textit{induces} between diseases.
Some extractions were correct, but many extractions mixed up the relation's direction (a causes b instead of b causes a).
In the end, PathIE was not very helpful for extracting such directed relations due to its poor quality.
We stopped the cleaning here, but a more advanced cleaning would be helpful to handle such situations.

\textit{Application Costs.}
We spent most of our time designing entity and relation vocabularies and analyzing the retrieved results. 
The creation of suitable vocabularies took as around one week in sum.
The execution of the toolbox scripts was quite simple; See our GitHub Repository.
To measure the runtime for PubPharm, we applied the PathIE-based pipeline on around 12 million PubMed abstracts (PubMed subset about drugs). 
The procedure could be completed within one week: 
Entity detection took two days for the complete PubMed collection (33 million abstracts).
PathIE took five days and cleaning took one day.
Hence, such an extraction workflow is realizable for PubPharm with moderate costs.

\textit{Generalizability.}
We already know that \openie{} and PathIE have worse performance than supervised methods; See the benchmarks in the original toolbox paper.
However we could design a suitable extraction workflow with an acceptable amount of time (a few weeks of cooperation with nine sessions with experts).
\openieS{} had a very poor recall, and filtering remained unclear.
Thus, they were not of interest for PubPharm's purposes.

\begin{table*}
  \centering
  \caption{Pollux \openieS{} example extractions. On the left the corresponding entity filter is shown (subject, partial and exact). }
  \label{tab:pollux_extractions}
  \begin{tabular}{@{}|l|l|lL{0.8\textwidth}|@{}}
     \toprule
         \multirow{7}{*}{\rotatebox[origin=c]{90}{\textbf{Political Sciences}}} &
         \parbox[t]{2mm}{\multirow{4}{*}{\rotatebox[origin=c]{90}{\textbf{Partial}}}}&
        PS1.1 & Stalin wanted all 16 \subjectAnnotated{Soviet}{NORP} Republics \predicate{to have} separate seats in \objectAnnotated{UN General Assembly}{ORG} but only 3 were given Russia Ukraine Belarus.\\
        && PS1.2 & This paper seeks to understand why the \subjectAnnotated{United States}{GPE} \predicate{treated} \objectAnnotated{Japan}{GPE} and Korea \predicate{differently} in the revisions of bilateral nuclear cooperation agreements.\\
         \cmidrule{2-4}
         & \parbox[t]{2mm}{\multirow{3}{*}{\rotatebox[origin=c]{90}{\textbf{Subject}}}}&
         PS2.1 & Based on these features, the article suggests that \subjectAnnotated{China}{GPE} is poised \predicate{to become} \object{a true global power}.\\
         && PS2.2 & Prior to the introduction of the Transparency Register the \subjectAnnotated{European Parliament}{ORG} \predicate{had maintained} \object{a Register of Accredited Lobbyists since 1996} while the European Commission [...].\\
     \bottomrule
  \end{tabular}
\end{table*}

PubPharm is currently using the PathIE extractions in their narrative retrieval service~\cite{kroll2021narrativequerygraphs}. 
Here recall is essential to find a suitable number of results to answer queries.
Although the quality of PathIE is only moderate, the quality seems to be sufficient for such a retrieval service.  
Here the statement should hint that the searched information is expressed within the document, e.g., that a \textit{metformin treatment} is contained. 
The main advantage of a retrieval service is that the original sentences can be shown to users to explain where the statements were extracted. 
In summary, if users are integrated into the process, and the statements'  origin is shown, these PathIE allow novel applications like the retrieval service.

Nevertheless, we encounter several issues: 
First, PathIE extracts wrong statements if several entities are contained in a sentence.
Next the undirected extractions of PathIE are often problematic if no additional cleaning can be performed (e.g., relations between diseases). 
Although these issues must be faced somehow, PathIE allowed us an extraction workflow that we could not have realized using supervised methods due to the lack of training data.
We would not recommend PathIE for building a knowledge graph due to many wrong extractions that would lead to transitive errors when performing reasoning on the resulting graph.

\textit{What is missing?}
In this pharmaceutical case study we focused on relations between pharmaceutical entities. 
PathIE completely ignored the surrounding context of statements, e.g., dose and duration information of therapies.
The coherence of statements was also broken down, e.g., drug, dosage form, disease, and target group of treatments were split into four separate statements. 
The desired goal would be to retain all relevant information within a single statement.
However PathIE is restricted to binary relations.
A future enhancement of PathIE would be desirable to retain all connected entities in a sentence.  
PubPharm's retrieval service bypassed the problem by using document contexts, i.e., statements from the same document belong together. 
The service uses abstracts, and this approximation would not have been possible for full texts because a full-text document might contain several different contexts.

\subsection{Political Sciences}
We applied the toolbox to 10k abstracts from political sciences.

\textit{Entity linking.}
The field of political sciences displays some distinct differences compared to the biomedical field and encyclopedias like Wikipedia. 
A notable difficulty lies in the lack of well-curated vocabularies for the domain. 
This can be mitigated in two ways: by using NER as implemented by Stanza~\cite{qi2020stanza} or by constructing/deriving entity vocabularies from general-purpose knowledge bases like Wikidata.
We investigated both approaches.

Stanza NER yielded ca. eight tags per document. 
The extracted mentions seemed sensible, e.g., entities like \textit{USA}, \textit{Bush} or \textit{the Cold War} were extracted. 
However Stanza NER also displayed some drawbacks, e.g., it was sensitive to missing uppercase letters for identifying names.
Such restrictions can be problematic in practice due to bad metadata (abstracts in upper case).

For the second approach we selected wars (Q198), coup d'états (Q45382) and elections (Q40231) as seed events, since those are likely to be subject of debate in political science articles.
Furthermore we inductively utilized Wikidata's subclass property (P279) to receive all subclasses of all seed events.
We used the SPARQL endpoint to export the corresponding vocabularies by asking for the English label and alias labels for the seed events, all instances of the seed events, and their subclasses.
In total, we collected 2.9k wars, 904 coups, and 79.7k election entries.
An evaluation of the toolbox's entity linker showed good performance on wars while coup d'états and elections were rarely linked sensible.
However we increased the linking quality by applying simple rules, e.g., the entity label must contain the term \textit{election}.
We derived 3.7k entities in sum.

\textit{Information Extraction.}
Due to the lack of comprehensive entity vocabularies, we focused on \openieS{} in this case study and omitted PathIE. 
\openieS{} yielded 147.2k (no filter), 28.6k (partial), 128 (exact) and 7.3k (subject) extractions.
Subject phrases tended to be short (only 32.0\% were complex), and object phrases tended to be long (74.3\% complex) again, like in the previous case studies. 
93.2\% of all sentences were estimated to be complex.
We randomly sampled 100 extractions of each filter for further analysis.
Again, extractions from small sentences looked helpful, while long sentences led to long object phrases.
We picked some interesting results and display them in Table~\ref{tab:pollux_extractions}.

\textit{Exact entity filter.}
Again the exact entity filter decreased the number of extractions drastically (from 147.2k to 128). 
But extractions seemed plausible, e.g., \textit{Alexander Lukashenko is president of Belarussian[SIC]} from \textit{Focus on the career and policies of the first Belarussian president, Alexander Lukashenko, elected in 1994.}
Another correct extraction was \textit{United States prepares to exit} from \textit{As the United States prepares to exit Afghanistan [...]}.

\textit{Partial entity filter.} In PS1.1, the extraction \textit{Soviet to have UN General Assembly} was wrong because the context about Stalin and separate seats was missed.
The extraction in PS1.2,  \textit{United States treated differently Japan}, was not helpful because \textit{Korea} was missed. 
Again, the context that this statement was investigated in that article was lost.
We found the extractions of the partial filter not helpful: 
Either they mixed up the original information or decisive context was missed.

\textit{Subject entity filter.}
The extraction PS2.1 showed a correct extraction, but then the information that the statement was suggested by an article was missed.
Although the sentence of P2.2 was quite complex, \openieS{} extracted useful information about the \textit{European Parliament}: \textit{European Parliament had maintained a Register of Accredited Lobbyists since 1996}.

We skipped the canonicalization procedure here because we already knew that canonicalizing \openieS{} verb phrases remains unclear (see Wikipedia case study).  
The exact filter yielded fewer extractions, partial filtering resulted in incorrect statements, and PathIE could not be applied due to the lack of vocabularies. 
And extractions from the subject filter could hardly be canonicalized to precise relations if the object phrase contains large sentence parts.

\textit{Application Costs.}
The application costs for the political domain seemed higher compared to the other two case studies. 
The lack of curated vocabularies necessitates the creation of such. 
As demonstrated, this can hardly be done automatically but requires domain knowledge. 
We exported some vocabularies from Wikidata but we missed many entities in the end.
In sum, we had four sessions, each 1.5 hours, with a domain expert to analyze the results.
The case study took us five person-days in sum.

\textit{Generalizability.} 
Due to the lack of available benchmarks, we restricted our evaluation to a qualitative level.
As another difficulty, simple fact statements, e.g., \textit{Joe Biden is the president of the USA} hardly carried new or relevant information. 
Still disputed claims, viewpoints, or assessments like \textit{the UK aims to position itself as an independent power after Brexit} might be the subject of study. 
This often resulted in long clauses for the subjects and objects that are hard to map to the already sparsely recognized named entities. 
But the subject entity filter allowed us to retain that \textit{UK aims to position itself as an independent power after Brexit} as a suitable extraction.
We plan to proceed from here by extracting semi-structured information via the subject filter.

\textit{What's Missing.}
Additionally the context of a statement is often highly relevant. 
In the example the statement loses its information if the context \textit{after Brexit} is omitted. 
Observations were similar to the Wikipedia case studies: 
Either the object phrases retained the context but could hardly be handled by filtering methods.
Or the object phrases were short and missed information.

\section{Discussion}

\begin{table*}
    \centering
    \caption{The table summarizes the measured runtimes for the samples and gives an estimation for the whole collection.}
    \label{tab:execution_remarks}
    \begin{tabular}{@{} |ll|cc|cc|cc| @{}}
        \toprule
          &&  \multicolumn{2}{c|}{Wikipedia} & \multicolumn{2}{c|}{Pharmacy} & \multicolumn{2}{c|}{Political Sciences} \\
          && \emph{Sample} & \emph{Estimation} &  \emph{Sample} & \emph{Estimation} &  \emph{Sample} & \emph{Estimation}    \\
         \midrule
           \multirow{2}{*}{Entity Det.} & \emph{NER} & 10.5 min & 19.4 days& - & - & 10.1 min & 21.6 hours \\
           & \emph{EL} &0.6 min & 1.2 days & 1.2 min & 2.8 days  & 0.7 min & 1.4 hours \\
           \midrule
           \multirow{2}{*}{Extraction} & \emph{PathIE} &2.6 min & 4.7 days&  2.0 min & 4.6 days & - & - \\
           & \emph{\openieS} & 53.6 min & 98.8 days&  74.0 min & 98.8 days  & 55.4 min & 5.0 days \\
           \midrule
           Cleaning & & < 1 hour & <1 day  & < 1 hour & <1 day & < 1 hour & < 1 day  \\
        \bottomrule
    \end{tabular}
\end{table*}

In the following we discuss how suitable unsupervised extraction workflows are in digital libraries by considering technical and conceptual limitations.
Furthermore we give best practices on what to do and when supervision is necessary.

\subsection{Technical Toolbox Limitations}
The toolbox filtered verb phrases by removing non-verbs (stop words, adverbs, etc.) and verbs like \textit{be} and \textit{have}.
Here negations in verb phrases were lost, too.
We implemented a parameter to make this behavior optional.
Next we implemented the subject entity filter that was useful in Wikipedia and political sciences.
Here a statement's subject must be linked to an entity, but the object can keep the original information. 
Then the results could be used as a semi-structured knowledge base, e.g., showing all actions of \textit{Albert Einstein} or \textit{positions} that the \textit{EU} has taken.

In addition, the dictionary-based entity linker fails to resolve short and ambiguous mentions. 
These wrongly linked mentions cause problems in the cleaning step (entity-based filters).
Here more advanced linkers would be more appropriate to improve the overall quality.
A coreference resolution is also missing, i.e., resolving all pronouns and mentions that refer to known entities.

PathIE is currently restricted to binary relations but might be extended to extract more higher-ary relations, e.g., by considering all connected entities via a verb phrase or a particular keyword like treatment.
A suitable cleaning would be possible if the relation arguments could be restricted to entity types. 

\subsection{Restrictions of Unsupervised Extraction}
The first significant restriction of unsupervised methods is their focus on and thus restriction to grammatical structures.
Suppose the example: \textit{The German book Känguru-Chroniken was written by Marc-Uwe Kling}. 
Here unsupervised methods may not extract that the language of the work is German.

In common relation extraction benchmarks such relations do appear and can be learned and inferred by modern language models~\cite{jinhyuk2019biobert,devlin2019bert}. 
However we argue that such extractions require high domain knowledge, typically unavailable in unsupervised extraction methods.
Similar examples could be made in specialized domains like pharmacy (treatments, inhibitions, etc.).
Moreover it is not possible to integrate this knowledge into unsupervised models by design: 
The model would need training data to infer such rules and, thus, be supervised.
We do not expect unsupervised models with access to comprehensive domain-specific knowledge soon.

Our case studies showed that \openieS{} extracts noun phrases in two ways: 
Either noun phrases are short and miss relevant information from the sentence.
These phrases are easier to handle but may be unhelpful in the end.
Or the noun phrases are long and complex but retain the original information.
Handling complex phrases requires more advanced cleaning methods.

The toolbox canonicalization procedure for relations considers only the verb phrases, not the surrounding context.
Verb phrases like \textit{uses}, \textit{publish}, and \textit{prevent} could refer to a plethora of relations.
In the end more advanced methods are required for a suitable canonicalization quality.
Especially canonicalizing \openieS{} verb phrases to precise relations was not really possible.

\subsection{Application and Costs}
Although we observed several issues and limitations, these methods can be used to implement services in digital libraries.
We summarize the measured runtimes and computed estimations for the corresponding collections in Table~\ref{tab:execution_remarks}.

Consider PubPharm for a good example: 
PathIE could enable a graph-based retrieval service with moderate costs~\cite{kroll2021narrativequerygraphs}. 
Around nine sessions with experts and moderate development time were necessary to implement a workflow.
The computation of PathIE took 2 min on our sample and was estimated to take 4.6 days for the whole PubMed collection.
Indeed, PubPharm could perform the complete extraction workflow in one week.

Our current cooperation with Pollux revealed that \openieS{} could bring more structure in this domain.
We will continue our work with Pollux by focusing on research questions that we would like to answer with semi-structured information derived from \openieS{} with subject entity filtering.

On our server with an Nvidia GTX 1080 TI, the computation of \openieS{}  took 55.4 min on the Pollux sample and is estimated to take five days for the complete collection.
For Wikipedia the sample took 53.6 min, and all English articles would require 98.8 days.
Note that we used a single GPU which is already five years old.
Hence the workflow can be accelerated with a modern GPU and parallelized by utilizing multiple GPUs.
In addition, \openieS{} can also be restricted to sentences that contain at least two entities.
Here the runtime is decreased from 55.4 to 22.4 min (Pollux) and 53.6 to 41.4 min (Wikipedia).

\subsection{Best Practices}
Subsequently we give some advice that we can deduce from our case studies.
\openieS{} handles short and simple sentences well. 
Here the exact entity filter will produce suitable extractions but decrease the recall drastically. 
The partial entity filter improves the recall but often messes up the original information. 
We recommend two strategies for long and complex sentences:

First, do not use the exact or partial entity filter because important information can be missed.
Use the subject entity filter to retrieve precise entities as subjects and the original information in object phrases. 
This filter allows the construction of semi-structured knowledge bases, e.g., positions that were taken by the \textit{EU} or actions that \textit{Albert Einstein} has done. 
Another option is to use no filter, but then, the extractions are not cleaned in any way.

Second, PathIE can find specialized relations that are expressed by keywords.
But PathIE requires directed relations that must be cleaned by entity type constraints.
Detecting such relations via PathIE is fast and probably cheaper than training supervised extraction models.
However PathIE will fail if several entities of the same type are mentioned within a sentence, e.g., side effects of treatments. 
Here supervised methods are required to achieve suitable quality.

\section{Related Work}
The main goal of information extraction (IE) is the extraction of structured information from unstructured or semi-structured information such as texts, tables, figures, and more~\cite{weikum2021machine,liu2007tableseer,manning2014stanford,kolluru2020openie6}.
In the following we give an overview of challenges and research trends in IE from texts.

\textit{Current Trends.} Modern IE research mainly focuses on improving the extraction accuracy, which is typically measured on benchmarks~\cite{kolluru2020openie6,bhardwaj2019carb}. 
Indeed, previous evaluations have shown that IE methods already produce good results, but the research is still ongoing~\cite{kolluru2020openie6,bhardwaj2019carb,devlin2019bert,niklaus2018iesurvey,kroll2021toolbox}. 
Primarily driven by the development of modern language models like BERT~\cite{devlin2019bert}, IE has made a huge step forward.

However these systems rely on supervised learning and thus need large-scale training data that cannot be reliably transferred across domains.
In brief, although supervised methods are up to the job with reasonable quality, their practical application comes at high costs. 
The expenses for supervision lead to the design of zero-shot, semi-supervised, and distant supervised extraction methods (see \cite{weikum2021machine} for a good overview).

\textit{Open Information Extraction.}  
Instead of designing extraction systems for each domain, methods like unsupervised information extraction (\openie) are proposed to change the game~\cite{niklaus2018iesurvey}.
\openie{} aims to extract knowledge from texts without knowing the entity and relation domains a-priori~\cite{weikum2021machine,niklaus2018iesurvey}. 
While supervised (closed) methods  focus on domain-specific and relevant relations and concepts, open methods are more flexible and may be applied across domains~\cite{weikum2021machine,niklaus2018iesurvey}. 
Vashishth proposed CESI to canonicalize \openie{} extractions by clustering noun and verb phases with the help of side information~\cite{vashishth2018cesi}.
However CESI was analyzed for short phrases that refer to precise entities. 
In addition studies have shown that \openie{} methods may struggle to handle scientific texts well because sentences are often long and domain-specific vocabulary terms are used~\cite{groth2018openiescientifictext}. 
While research in both directions (open and closed) is still ongoing, some works bridge the gap between both worlds:
Kruiper et al. propose the task of Semi-Open Relation extraction~\cite{kruiper2020semiopenrelationextraction}, i.e., they use domain-specific information to filter irrelevant open information extractions. 
Similarly, we showed that domain-specific filtering of \openie{} outputs could yield helpful results~\cite{kroll2021toolbox}.

\textit{Information Extraction in Digital Libraries.}
Digital libraries are interested in practical IE workflows to allow novel applications; See this tutorial at JCDL2016~\cite{kyle2016jcdltutorialinformationextraction}. 
IE can allow literature-based discovery workflows, which have been studied on DBpedia~\cite{menasha2020jcdllbdworkflow}.
The extraction of entities and relations is therefore challenging. 
That is why modern approaches build upon language models and supervision for a reliable extraction~\cite{sai2021eekejcdlentityandrelationextraction}.
These language models require extensive computational resources for training and application~\cite{devlin2019bert,jinhyuk2019biobert}.
Good examples for IE are DBpedia~\cite{auer2007dbpedia} that was harvested from Wikipedia infoboxes or the SemMedDB, which is a collection of biomedical statements harvested from PubMed~\cite{zhang2014ddisemmeddb,kilicoglu2012semmeddb}. 
Hristovski et al. have used the SemMedDB to perform knowledge discovery~\cite{hristovski2015semmeddblbd}.
Nevertheless the construction of SemMedDB required biomedical experiences to define hand-written rules for the extraction.
In contrast to the previous works, our work focused on nearly-unsupervised extraction workflows that do not rely on training data for the extraction phase.

\section{Conclusion}
In this paper we have studied nearly-unsupervised extraction workflows for a practical application in digital libraries.
We focused on three different domains to generalize our findings, namely the encyclopedia Wikipedia, pharmacy, and political sciences.
First, the scalability of the investigated methods was acceptable for our partners.
Second, unsupervised extraction workflows required intensive cleaning and canonicalization to result in precise semantics.
Thus they do not work out-of-the-box and reliably canonicalize \openie{} verb phrases remains an open issue.
Although such cleaning can be exhausting, the pharmaceutical case study yielded a novel retrieval service.
Such a service would not have been possible when training data must have been collected for each relation.
In addition, not filtering complex object phrases can allow the construction of semi-structured knowledge bases or enrich the original texts, e.g., show all actions of Albert Einstein.
In conclusion, unsupervised extraction workflows are worth studying in digital libraries.
They come with limitations and require cleaning, but they entirely bypass the lack of training data in the extraction phase.

\section*{Acknowledgment}
Supported by the Deutsche Forschungsgemeinschaft (DFG, German Research Foundation): PubPharm – the Specialized Information Service for Pharmacy (Gepris 267140244).

\bibliographystyle{ACM-Reference-Format}
\bibliography{references.bib}

\end{document}